\documentclass[10pt,twocolumn,letterpaper]{article}

\usepackage{cvpr}
\usepackage{times}
\usepackage{epsfig}
\usepackage{graphicx}
\usepackage{amsmath}
\usepackage{amssymb}
\usepackage{comment}
\usepackage{multirow}
\usepackage{subfigure}
\usepackage{caption}
\usepackage{algorithm}
\usepackage{algorithmic}
\usepackage{booktabs}
\usepackage{multirow}
\usepackage{bigstrut}
\usepackage{url}

\usepackage[pagebackref=true,breaklinks=true,letterpaper=true,colorlinks,bookmarks=false]{hyperref}

\cvprfinalcopy 

\def\cvprPaperID{3269} 

\ifcvprfinal\fi
\begin{document}

\title{Unsupervised Person Image Generation with Semantic Parsing Transformation}

\author{Sijie Song{$\small^{1}$}, Wei Zhang{$\small^{2}$}, Jiaying Liu{$\small^{1}$}\thanks{Corresponding author. This work was done at JD AI Research. Our project is available at \url{https://github.com/SijieSong/person_generation_spt.git}.}, Tao Mei{$\small^{2}$}\\
$\small ^{1}$\ Institute of Computer Science and Technology, Peking University, Beijing, China\\
$\small ^{2}$\ JD AI Research, Beijing, China
}
\maketitle

\begin{abstract}
\vspace{-4mm}

In this paper, we address unsupervised pose-guided person image generation, which is known challenging due to non-rigid deformation. Unlike previous methods learning a rock-hard direct mapping between human bodies, we propose a new pathway to decompose the hard mapping into two more accessible subtasks, namely, semantic parsing transformation and appearance generation.
Firstly, a semantic generative network is proposed to transform between semantic parsing maps, in order to simplify the non-rigid deformation learning. Secondly, an appearance generative network learns to synthesize semantic-aware textures. Thirdly, we demonstrate that training our framework in an end-to-end manner further refines the semantic maps and final results accordingly. Our method is generalizable to other semantic-aware person image generation tasks, \eg, clothing texture transfer and controlled image manipulation. Experimental results demonstrate the superiority of our method on DeepFashion and Market-1501 datasets, especially in keeping the clothing attributes and better body shapes.

\end{abstract}

\vspace{-3mm}
\section{Introduction}

Pose-guided image generation has attracted great attentions recently, which is to change the pose of the person image to a target pose, while keeping the appearance details. This topic is of great importance in fashion and art domains for a wide range of applications from image / video editing, person re-identification to movie production.



\begin{figure}[t] 
	\begin{center}
		\includegraphics[width=1.0\linewidth]{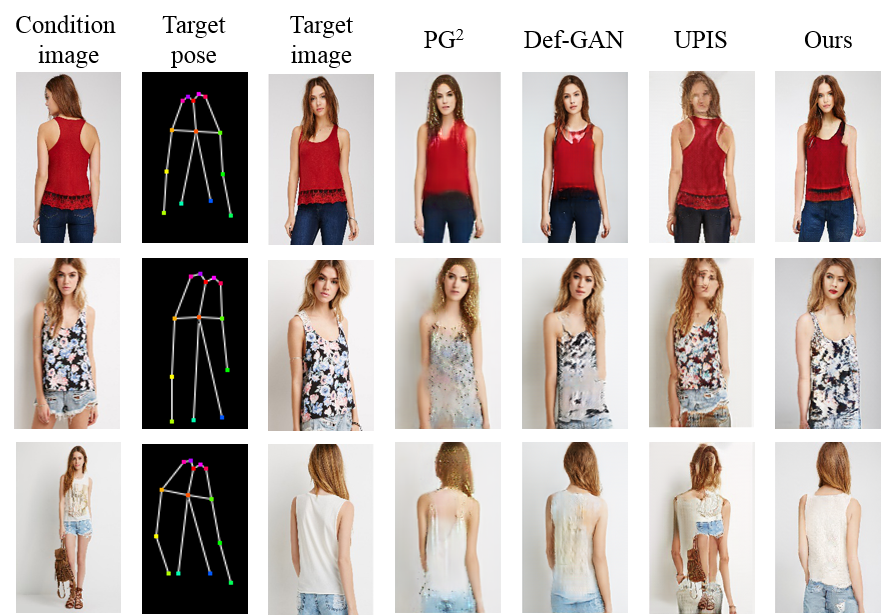}
	\end{center}
\vspace{-5mm}
	\caption{Visual results of different methods on DeepFashion~\cite{liu2016deepfashion}. Compared with PG$^2$~\cite{ma2017pose}, Def-GAN~\cite{siarohin2018deformable}, and UPIS~\cite{pumarola2018unsupervised}, our method successfully keeps the clothing attributes (\emph{e.g.}, textures) and generates better body shapes (\emph{e.g.}, arms).}
\vspace{-5mm}
	\label{fig:teaser}
\end{figure}


With the development of deep learning and generative model~\cite{goodfellow2014generative}, many researches have been devoted to pose-guided image generation~\cite{ma2017pose,pumarola2018unsupervised,esser2018variational,siarohin2018deformable,si2018multistage,balakrishnan2018synthesizing,ma2017disentangled}. Initially, this problem is explored under the fully supervised setting~\cite{ma2017pose,siarohin2018deformable,si2018multistage,balakrishnan2018synthesizing}. Though promising results have been presented, their training data has to be composed with paired images (\emph{i.e.}, same person in the same clothing but in different poses).
To tackle this data limitation and enable more flexible generation, more recent efforts have been devoted to learning the mapping with unpaired data~\cite{pumarola2018unsupervised,esser2018variational,ma2017disentangled}. However without ``paired" supervision, results in~\cite{pumarola2018unsupervised} are far from satisfactory due to the lack of supervision.
Disentangling image into multiple factors (\eg, background / foreground, shape / appearance) is explored in~\cite{ma2017disentangled,esser2018variational}. But ignoring the non-rigid human-body deformation and clothing shapes leads to compromised generation quality.



Formally, the key challenges of this unsupervised task are in three folds. First, due to the non-rigid nature of human body, transforming the spatially misaligned body-parts is difficult for current convolution-based networks. Second, clothing attributes, \eg, sleeve lengths and textures, are generally difficult to preserve during generation. However, these clothing attributes are crucial for human visual perception. Third, the lack of paired training data gives little clue in establishing effective training objectives.

To address these aforementioned challenges, we propose to seek a new pathway for unsupervised person image generation. Specifically, instead of directly transforming the person image, we propose to transform the semantic parsing between poses. On one hand, translating between \emph{person image} and \emph{semantic parsing} (in both directions) has been extensively studied, where sophisticated models are available. On the other hand, semantic parsing transformation is a much easier problem to handle spatial deformation, since the network does not care about the appearance and textures.



As illustrated in Fig.~\ref{fig:framework}, our model for unsupervised person image generation consists of two modules: semantic parsing transformation and appearance generation. In semantic parsing transformation, a semantic generative network is employed to transform the input semantic parsing to the target parsing, according to the target pose. Then an appearance generative network is designed to synthesize textures on the transformed parsing. Without paired supervision, we create pseudo labels for semantic parsing transformation and introduce cycle consistency for training. Besides, a semantic-aware style loss is developed to help the appearance generative network learn the essential mapping between corresponding semantic areas, where clothing attributes can be well-preserved by rich semantic parsing. Furthermore, we demonstrate that the two modules can be trained in an end-to-end manner for finer semantic parsing as well as the final results.

In addition, the mapping between corresponding semantic areas inspires us to apply our appearance generative network on applications of semantic-guided image generation. Conditioning on the semantic map, we are able to achieve clothing texture transfer of two person images. In the meanwhile, we are able to control the image generation by manually modifying the semantic map.

The main contributions can be summarized as follows:
\begin{itemize}
    \item We propose to address the unsupervised person image generation problem. Consequently, the problem is decomposed into semantic parsing transformation ($H_S$) and appearance generation ($H_A$).

    \item We design a delicate training schema to carefully optimize $H_S$ and $H_A$ in an end-to-end manner, which generates better semantic maps and further improves the pose-guided image generation results.

    \item Our model is superior in rendering better body shape and keeping clothing attributes. Also it is generalizable to other conditional image generation tasks, \eg, clothing texture transfer and controlled image manipulation.
\end{itemize}

\section{Related Work}
\subsection{Image Generation}
With the advances of generative adversarial networks (GANs)~\cite{goodfellow2014generative}, image generation has received a lot of attentions and been applied on many areas~\cite{karras2017progressive,wang2018pix2pixHD,StarGAN2018,xian2017texturegan}. There are mainly two branches in this research field. One lies in supervised methods and another lies in unsupervised methods. Under the supervised setting, pix2pix~\cite{isola2017image} built a conditional GAN for image to image translation, which is essentially a domain transfer problem. Recently, more efforts~\cite{karras2017progressive,wang2018pix2pixHD} have been devoted to generating really high-resolution photo-realistic images by progressively generating multi-scale images. For the unsupervised setting, reconstruction consistency is employed to learn cross-domain mapping~\cite{zhu2017unpaired,yi2017dualgan,kim2017learning}. However, these unsupervised methods are developed and applied mostly for appearance generation of the spatially aligned tasks. With unpaired training data, our work is more intractable to learn the mapping to handle spatial non-rigid deformation and appearance generation simultaneously.


\subsection{Pose-Guided Person Image Generation}
The early attempt on pose-guided image generation was achieved by a two-stage network PG$^2$~\cite{ma2017pose}, in which the output under the target pose is coarsely generated in the first stage, and then refined in the second stage. To better model shape and appearance, Siarohin \emph{et al.}~\cite{siarohin2018deformable} utilized deformable skips to transform high-level features of each body part. Similarly, the work in~\cite{balakrishnan2018synthesizing} employs body part segmentation masks to guide the image generation. However,~\cite{ma2017pose,siarohin2018deformable,balakrishnan2018synthesizing} are trained with paired data.
To relieve the limitation, Pumarola \emph{et al.}~\cite{pumarola2018unsupervised} proposed a fully unsupervised GAN, borrowing the ideas from~\cite{zhu2017unpaired,reed2016learning}. On the other hand, the works in~\cite{esser2018variational,ma2017disentangled} solved the unsupervised problem by sampling from feature spaces according to the data distribution. These sample based methods are less faithful to the appearance of reference images, since they generate results from highly compressed features. Instead, we use semantic information to help preserve body shape and texture synthesis between corresponding semantic areas.

\subsection{Semantic Parsing for Image Generation}
The idea of inferring scene layout (semantic map) has been explored in~\cite{hong2018inferring,johnson2018image} for text-to-image translation. Both of the works illustrate that by conditioning on estimated layout, more semantically meaningful images can be generated. The scene layout is predicted from texts~\cite{hong2018inferring} or scene graphs~\cite{johnson2018image} with the supervision from groundtruth. In contrast, our model learns the prediction for semantic map in an unsupervised manner.
We also show that the semantic map prediction can be further refined by end-to-end training.

\begin{figure*}[t] 
	\begin{center}
		\includegraphics[width=0.8\linewidth]{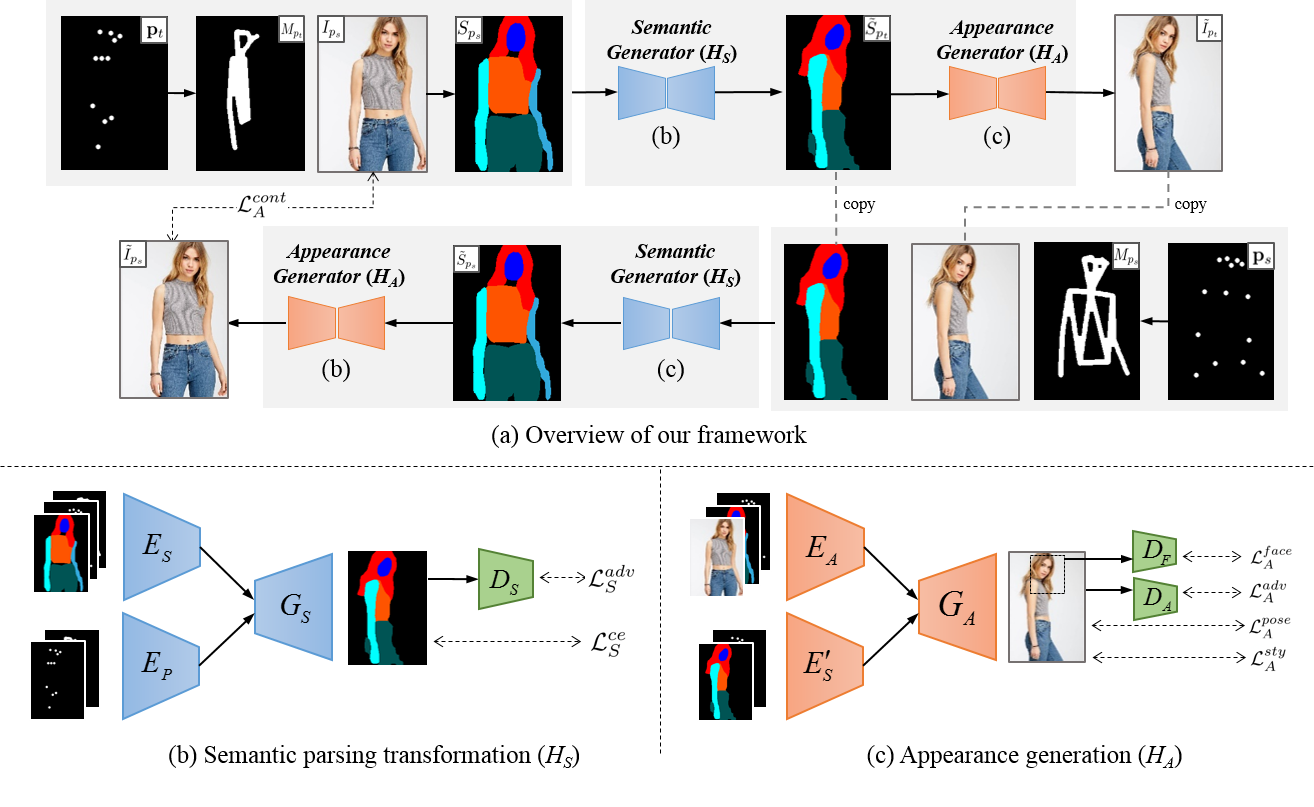}
	\end{center}
\vspace{-5mm}
	\caption{Our framework for unsupervised person image generation.}
\vspace{-5mm}
	\label{fig:framework}
\end{figure*}

\section{The Proposed Method}
Given a target pose $\mathbf{p}_t$ and a reference image $I_{p_s}$ under pose $\mathbf{p}_s$, our goal is to generate an output image $\tilde I_{p_t}$, which follows the clothing appearance of $I_{p_s}$ but under the pose $\mathbf{p}_t$. This generation can be formulated as: $<I_{p_s}, \mathbf{p}_t> \rightarrow \tilde I_{p_t}$.

During the training process, we are under an unsupervised setting: the training set is composed with $\{I^{i}_{p^i_s}, \mathbf{p}^i_s, \mathbf{p}^i_t\}_{i=1}^N$, where the corresponding ground-truth image $I^i_{p_t}$ is not available. For this challenging unpaired person image generation problem, our key idea is to introduce human semantic parsing to decompose it into two modules:\emph{ semantic parsing transformation} and \emph{appearance generation}. Our overall framework can be viewed in Fig.~\ref{fig:framework}(a). Semantic parsing transformation module aims to first generate a semantic map under the target pose, which provides crucial prior for the human body shape and clothing attributes. Guided by the predicted semantic map and the reference image, appearance generation module then synthesizes textures for the final output image.

In the following, we first introduce person representation, which is the input of our framework. We then describe each module in details from the perspective of independent training. Finally, we illustrate the joint learning of the two modules in an end-to-end manner.

\subsection{Person Representation}
Besides the reference image $I_{p_s} \in \mathbb{R}^{3\times H\times W}$, the source pose $\mathbf{p}_s$, and the target pose $\mathbf{p}_t$, our model also involves a semantic map $S_{p_s}$ extracted from $I_{p_s}$, pose masks ${M}_{p_s}$ for $\mathbf{p}_s$ and ${M}_{p_t}$ for $\mathbf{p}_t$. In our work, we represent poses as probability heat maps, \emph{i.e.}, $\mathbf{p}_s, \mathbf{p}_t \in \mathbb{R}^{k\times H \times W} (k=18)$. The semantic map $S_{p_s}$ is extracted with an off-the-shelf human parser~\cite{gong2017look}. We represent $S_{p_s}$ using a pixel-level one-hot encoding, \emph{i.e.}, $S_{p_s} \in \{0,1\}^{L\times H \times W}$, where $L$ indicates the total number of semantic labels. For the pose masks ${M}_{p_s}$ and $M_{p_t}$, we adopt the same definition in~\cite{ma2017pose}, which provide prior on pose joint connection in the generation process.

\subsection{Semantic Parsing Transformation ($H_S$)}

In this module, we aim to predict the semantic map $\tilde{S}_{p_t}\in [0,1]^{L\times H \times W}$ under the target pose $\mathbf{p}_t$, according to the reference semantic map $S_{p_s}$. It is achieved by the semantic generative network, which is based on U-Net~\cite{ronneberger2015u}. As shown in Fig.~\ref{fig:framework}(b), our semantic generative network consists of a semantic map encoder $E_S$, a pose encoder $E_P$ and a semantic map generator $G_S$.
$E_S$ takes $S_{p_s}$, $\mathbf{p}_s$ and ${M}_{p_s}$ as input to extract conditional sematic information, while $E_P$ takes $\mathbf{p}_t$ and ${M}_{p_t}$ as input to encode the target pose. $G_S$ then predicts $\tilde S_{p_t}$ based on the encoded features. As~\cite{zhu2017your}, \emph{softmax} activation function is employed at the end of $G_S$ to generate the semantic label for each pixel. Formally, the predicted semantic map $\tilde{S}_{p_t}$ conditioned on $S_{p_s}$ and $\mathbf{p}_t$ is formulated as $\tilde{S}_{p_t} = G_S\left(E_S(S_{p_s}, \mathbf{p}_s, {M}_{p_s}), E_P(\mathbf{p}_t,{M}_{p_t})\right)$. The introduction of $M_{p_s}$ and $ M_{p_t}$ as input is to help generate continuous semantic maps, especially for bending arms.


\textbf{Pseudo label generation}. The semantic generative network is trained to model the spatial semantic deformation under different poses. Since semantic maps do not associate with clothing textures, people in different clothing appearance may share similar semantic maps. Thus, we can search similar semantic map pairs in the training set to facilitate the training process. For a given $S_{p_s}$, we search a semantic map $S_{p^*_t}$ which is under different poses but shares the same clothing type as $S_{p_s}$. Then we use $\mathbf{p}^*_t$ as the target pose for $S_{p_s}$, and regard $S_{p^*_t}$ as the pseudo ground truth.
We define a simple yet effective metric for the search problem. The human body is decomposed into ten rigid body subparts as in~\cite{siarohin2018deformable}, which can be represented with a set of binary masks $\{B^j\}_{j=1}^{10}(B^j\in{\mathbb{R}^{H \times W}})$. $S_{p^*_t}$ is searched by solving
\vspace{-2mm}
\begin{equation}
\label{equ:parsing_metric}
S_{p^*_t} = \arg \min_{S_p} \sum_{j=1}^{10} || B^j_{p} \otimes S_{p} - f_j(B^j_{p_s} \otimes S_{p_s})||_2^2,
\vspace{-1mm}
\end{equation}
where $f_j(\cdot)$ is an affine transformation to align the two body parts according to four corners of corresponding binary masks, $\otimes$ denotes the element-wise multiplication. Note that pairs sharing very similar poses are excluded.

\textbf{Cross entropy loss}. The semantic generative networks can be trained under supervision with paired data $\{S_{p_s}, \mathbf{p}_s, S_{p^*_t}, \mathbf{p}^*_t\}$. We use the cross-entropy loss $\mathcal{L}_S^{ce}$ to constrain pixel-level accuracy of semantic parsing transformation, and we give the human body more weight than the background with the pose mask ${M}_{p^*_t}$as
\vspace{-1mm}
\begin{equation}
\label{equ:parsing_ce_loss}
\mathcal{L}_S^{ce} =  -||S_{p^*_t}\otimes log (\tilde S_{p^*_t}) \otimes (1+{M}_{p^*_t})||_1.
\vspace{-1mm}
\end{equation}

\textbf{Adversarial loss}. We also employ an adversarial loss $\mathcal L_S^{adv}$ with a discriminator $D_S$ to help $G_S$ generate semantic maps of \emph{visual style} similar to the realistic ones.
\vspace{-1mm}
\begin{equation}
\label{equ:parsing_adv_loss}
\begin{aligned}
\mathcal{L}_S^{adv} = \mathcal L^{adv}(H_S, D_S, S_{p^*_t}, \tilde S_{p^*_t}),
\vspace{-1mm}
\end{aligned}
\end{equation}
where $H_S = G_S\circ(E_S, E_P)$, $\mathcal L^{adv}(G,D,X,Y) = \mathbb{E}_{X}[\log D(X))] + \mathbb{E}_{Y}[\log(1 - D(Y)]$ and $Y$ is associated with $G$.

The overall losses for our semantic generative network are as follows,
\vspace{-1mm}
\begin{equation}
\label{equ:parsing_total_loss}
\mathcal L_S^{total} = \mathcal{L}_S^{adv} + \lambda^{ce} \mathcal{L}_S^{ce}.
\vspace{-1mm}
\end{equation}

\subsection{Appearance Generation ($H_A$)}

In this module, we utilize the appearance generative network to synthesize textures for the output image $\tilde I_{p_t} \in \mathbb{R}^{3\times H \times W}$, guided by the reference image $S_{p_s}$ and predicted semantic map $\tilde S_{p_t}$ from semantic parsing transformation module. The architecture of appearance generative network consists of an appearance encoder $E_A$ to extract the appearance of reference image $I_{p_s}$, a semantic map encoder $E'_S$ to encode the predicted semantic map $\tilde S_{p_t}$, and an appearance generator $G_A$. The architecture of appearance generative network is similar to the semantic generative network, except that we employ deformable skips in~\cite{siarohin2018deformable} to better model spatial deformations. The output image is obtained by $\tilde I_{p_t} = G_A\left(E_A(I_{p_s}, S_{p_s}, \mathbf{p}_s), E'_S(\tilde S_{p_t},\mathbf{p}_t)\right)$, as in Fig.~\ref{fig:framework}(c).

Without the supervision of ground truth $I_{p_t}$, we train the appearance generative network using the cycle consistency as~\cite{zhu2017unpaired,pumarola2018unsupervised}, in which $G_A$ should be able to map back $I_{p_s}$ with the generated $\tilde I_{p_t}$ and $\mathbf{p}_s$. We denote the mapped-back image as $\tilde I_{p_s}$, and the predicted segmentation map as $\tilde S_{p_s}$ in the process of mapping back.

\textbf{Adversarial loss}.
Discriminator $D_A$ is first introduced to distinguish between the realistic image and generated image, which leads to adversarial loss $\mathcal{L}_A^{adv}$
\begin{equation}
\label{equ:image_adv_loss}
\begin{aligned}
& \mathcal{L}_A^{adv} = \mathcal L^{adv}(H_A,D_A,I_{p_s},\tilde I_{p_t}) + \mathcal L^{adv}(H_A,D_A,I_{p_s}, \tilde I_{p_s}),
\end{aligned}
\end{equation}
where $H_A = G_A\circ(E_A, E'_S)$.

\textbf{Pose loss}.
As in~\cite{pumarola2018unsupervised}, we use pose loss $\mathcal{L}^{pose}_A$ with a pose detector $\mathcal{P}$ to generate images faithful to the target pose
\vspace{-1mm}
\begin{equation}
\mathcal{L}_A^{pose} = ||\mathcal{P}(\tilde I_{p_t}) - \mathbf{p}_t||^2_2 +||\mathcal{P}(\tilde I_{p_s}) - \mathbf{p}_s||^2_2.
\label{equ:image_pose_det_loss}
\vspace{-1mm}
\end{equation}

\textbf{Content loss}.
Content loss $\mathcal{L}^{cont}_A$ is also employed to ensure the cycle consistency
\vspace{-1mm}
\begin{equation}
\label{equ:image_content_loss}
\mathcal{L}_A^{cont} = ||\Lambda(\tilde{I}_{p_s}) - \Lambda(I_{p_s})||_2^2,
\vspace{-1mm}
\end{equation}
where $\Lambda(I)$ is the feature map of image $I$ of \emph{conv2\_1} layer in VGG16 model~\cite{simonyan2015very} pretrained on ImageNet.

\textbf{Style loss}.
It is challenging to correctly transfer the color and textures from $I_{p_s}$ to $\tilde I_{p_t}$ without any constraints, since they are spatially misaligned. ~\cite{pumarola2018unsupervised} tried to tackle this issue with patch-style loss, which enforces that texture around corresponding pose joints in $I_{p_s}$ and $\tilde I_{p_t}$ are similar. We argue that patch-style loss is not powerful enough in two-folds: (1) textures around joints would change with different poses, (2) textures of main body parts are ignored. Another alternative is to utilize body part masks. However, they can not provide texture contour. Thanks to the guidance provided by semantic maps, we are able to well retain the style with a semantic-aware style loss to address the above issues. By enforcing the style consistency among $I_{p_s}$, $\tilde I_{p_t}$ and $\tilde I_{p_s}$, our semantic-aware style loss is defined as
\vspace{-1mm}
\begin{equation}
\label{equ:image_style_loss}
\mathcal{L}_A^{sty} = \mathcal L^{sty}(I_{p_s}, \tilde I_{p_t}, S_{p_s}, \tilde S_{p_t}) + \mathcal L^{sty}(\tilde I_{p_t}, \tilde I_{p_s}, \tilde S_{p_t}, \tilde S_{p_s}),
\vspace{-1mm}
\end{equation}
where
\begin{equation*}
\label{equ:style_loss}
\begin{aligned}
    &\mathcal{L}^{sty}(I_1, I_2, S_1, S_2) \\
 =  & \sum_{l=1}^{L} ||\mathcal G( \Lambda(I_1) \otimes \Psi_l(S_1)) - \mathcal G(\Lambda(I_2) \otimes \Psi_l(S_2)))||^2_2.
\end{aligned}
\vspace{-1mm}
\end{equation*}
And $\mathcal{G}(\cdot)$ denotes the function for Gram matrix~\cite{gatys2016image}, $\Psi_l(S)$ denotes the downsampled binary map from $S$, indicating pixels that belong to the $l$-th semantic label.

\textbf{Face loss}.
Besides, we add a discriminator $D_{F}$ for generating more natural faces,
\vspace{-1mm}
\begin{equation}
\label{equ:face_adv_loss}
\begin{aligned}
\vspace{-1mm}
 \mathcal{L}_A^{face} & = \mathcal L^{adv}(H_A,D_F, \mathcal F(I_{p_s}), \mathcal F(\tilde  I_{p_t})) \\
& + \mathcal L^{adv}(H_A,D_F,\mathcal F(I_{p_s}), \mathcal F(\tilde I_{p_s})),
\end{aligned}
\end{equation}
where $\mathcal{F}(I)$ represents the face extraction guided by pose joints on faces, which is achieved by a non-parametric spatial transform network~\cite{jaderberg2015spatial} in our experiments.

The overall losses for our appearance generative network are as follows,
\vspace{-1mm}
\begin{equation}
\label{equ:total_loss}
\begin{aligned}
\vspace{-1mm}
\mathcal{L}_A^{total} & = \mathcal L_A^{adv} + \lambda^{pose} \mathcal L_A^{pose} + \lambda^{cont} \mathcal L_A^{cont} \\
                      & + \lambda^{sty} \mathcal L_A^{sty} + \mathcal L_A^{face}.
\end{aligned}
\end{equation}

\subsection{End-to-End Training}
\begin{figure}[htbp] 
	\begin{center}
    \subfigure[Searching error]{
       \includegraphics[width=0.46\linewidth]{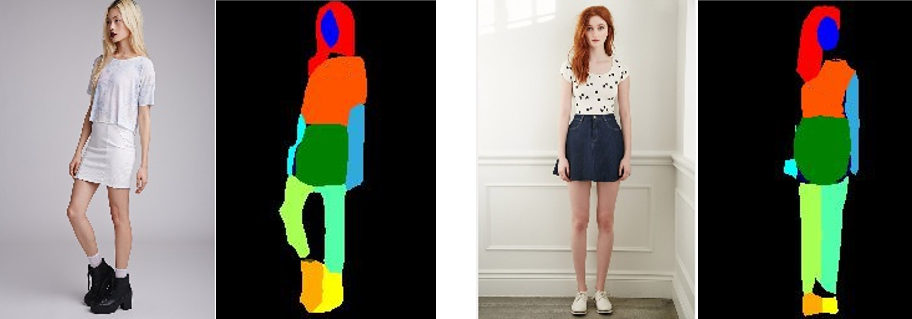}
        \label{fig:search_err}
    }
    \subfigure[Parsing error]{
       \includegraphics[width=0.46\linewidth]{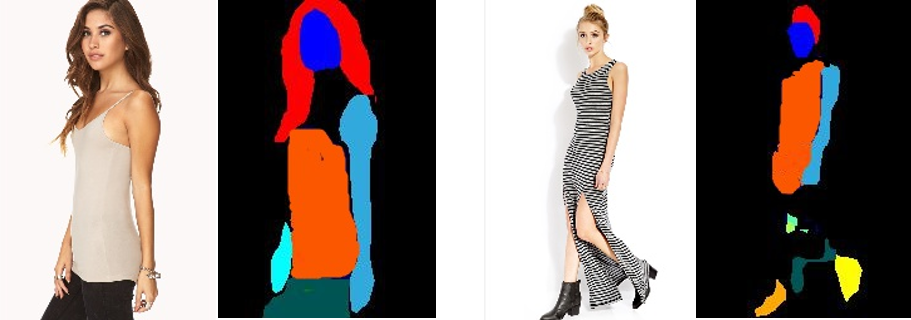}
        \label{fig:parsing_err}
    }
	\end{center}
\vspace{-6mm}
	\caption{Errors exist in the searched semantic map pairs, which might cause the inaccuracy of semantic parsing transformation.}
\vspace{-6mm}
	\label{fig:pseudo_label}
\end{figure}

Since the shape and contour of our final output is guided by the semantic map, the visual results of appearance generation rely heavily on the quality of predicted semantic map from semantic parsing transformation. However, if they are independently trained, two reasons might lead to instability for $H_S$ and $H_A$.
\vspace{-1mm}
\begin{itemize}
\vspace{-2mm}
  \item Searching error: the searched semantic maps are not very accurate, as in Fig.~\ref{fig:search_err}.
\vspace{-2mm}
  \item Parsing error: the semantic maps obtained from human parser are not accurate, since we do not have labels to finetune the human parser, as in Fig.~\ref{fig:parsing_err}.
\vspace{-2mm}
\end{itemize}


Our training scheme is shown in Algorithm~\ref{alg:e2e_training}.
\begin{algorithm}[htb]
\caption{End-to-end training for our network.}
\label{alg:e2e_training}
\begin{algorithmic}[1]
\REQUIRE $\{S^i_{p_s}, \mathbf{p}^i_s, S^i_{p^*_t},(\mathbf{p}^{*}_t)^i\}_{i=1}^{N^*}$, $\{I^i_{p_s}, \mathbf{p}^i_s, \mathbf{p}^i_t\}_{i=1}^N$.

\STATE Initialize the network parameters.

//Pre-train $H_S$

\STATE With $\{S^i_{p_s}, \mathbf{p}^i_s, S^i_{p^*_t}, (\mathbf{p}^{*}_t)^i\}_{i=1}^{N^*}$, train $\{H_S, D_S\}$ to optimize $\mathcal{L}_S^{total}$.

//Train $H_A$

\STATE With $\{I^i_{p_s}, \mathbf{p}^i_s, \tilde S^i_{p_t}, \mathbf{p}^i_t\}_{i=1}^N$ and $\{H_S, D_S\}$ fixed, train $\{H_A, D_A, D_{face}\}$ to optimize $\mathcal{L}_A^{total}$.

//Joint optimization
\STATE Train $\{H_S, D_S, H_A, D_A, D_{face}\}$ jointly with $\mathcal{L}_A^{total}$, using $\{I^i_{p_s}, \mathbf{p}^i_s, \tilde S^i_{p_t},\mathbf{p}^i_t\}_{i=1}^N$.

\ENSURE $H_S$, $H_A$.
\end{algorithmic}
\end{algorithm}

\section{Experiments}
\vspace{-2mm}
In this section, we evaluate our proposed framework with both qualitative and quantitative results.

 \begin{figure}[t] 
	\begin{center}
		\includegraphics[width=1.0\linewidth]{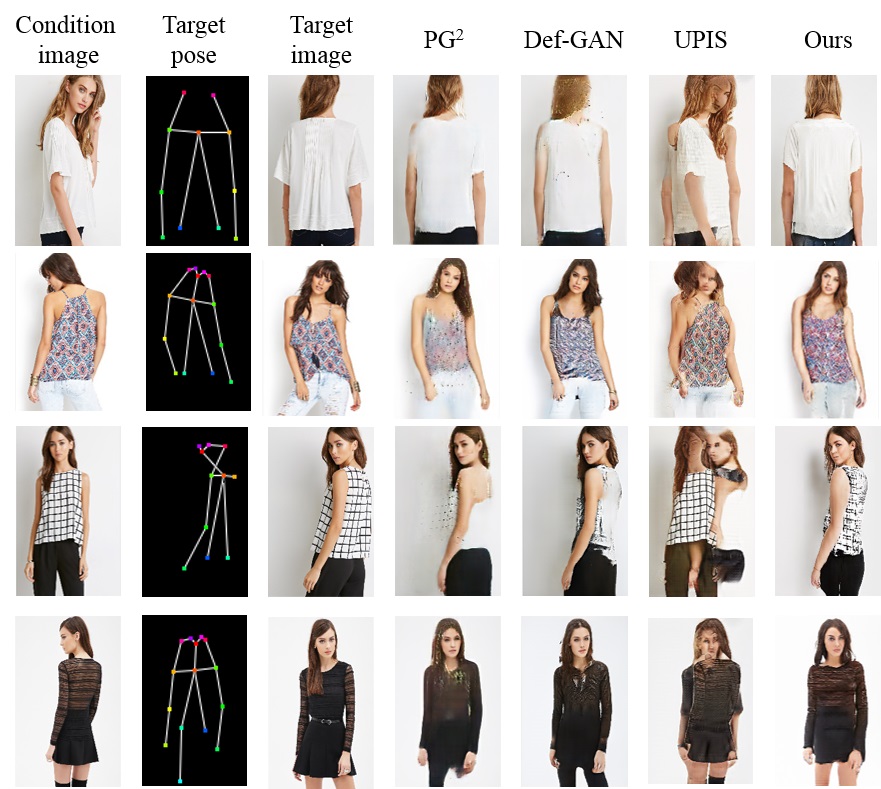}
	\end{center}
    \vspace{-4mm}
	\caption{Example results by different methods (PG${^2}$~\cite{ma2017pose}, Def-GAN~\cite{siarohin2018deformable} and UPIS~\cite{pumarola2018unsupervised}) on DeepFashion. Our model better keeps clothing attributes (\emph{e.g.}, textures, clothing types).}
    \vspace{-3mm}
	\label{fig:stoa_fashion}
\end{figure}

\subsection{Datasets and Settings}
\textbf{DeepFashion}~\cite{liu2016deepfashion}. We experiment with the \emph{In-shop Clothes Retrieval Benchmark} of the DeepFashion dataset. It contains a large number of clothing images with various appearance and poses, the resolution of which is $256~\times~256$. Since our method does not require paired data, we randomly select $37,258$ images for training and $12,000$ images for testing.

\textbf{Market-1501}~\cite{zheng2015scalable}. This dataset contains 32,668 images from different viewpoints. The images are in the resolution of $128\times 64$. We adopt the same protocol for data split as in~\cite{zheng2015scalable}. And we select 12,000 pairs for testing as in~\cite{siarohin2018deformable}.

\textbf{Implementation details.} For the person representation, the 2D poses are extracted using OpenPose~\cite{cao2017realtime}, and the condition semantic maps are extracted with the state-of-the-art human parser~\cite{gong2017look}. We integrate the semantic labels originally defined in~\cite{gong2017look} and set $L=10$ (\emph{i.e.}, background, face, hair, upper clothes, pants, skirt, left/right arm, left/right leg). For DeepFashion dataset, the joint learning to refine semantic map prediction is performed on the resolution of $128\times128$. Then we upsample the predicted semantic maps to train images in $256\times256$ with progressive training strategies~\cite{karras2017progressive}. For Market-1501, we directly train and test on $128\times 64$. Besides, since the images in Market-1501 are in low resolution and the face regions are blurry. $\mathcal L^{face}_A$ is not adopted on Market-1501 for efficiency. For the hyper-parameters, we set $\lambda^{pose}$, $\lambda^{cont}$  as 700, 0.03 for DeepFashion and 1, 0.003 for Market-1501. $\lambda^{sty}$ is 1 for all experiments. We adopt ADAM optimizer~\cite{kingma2014adam} to train our network with a learning rate 0.0002 ($\beta_1 = 0.5$ and $\beta_2=0.999$). The batch sizes for DeepFashion and Market-1501 are set to 4 and 16, respectively. For more detailed network architecture and training scheme on each dataset, please refer to our supplementary.

\begin{figure}[t] 
	\begin{center}
		\includegraphics[width=1.0\linewidth]{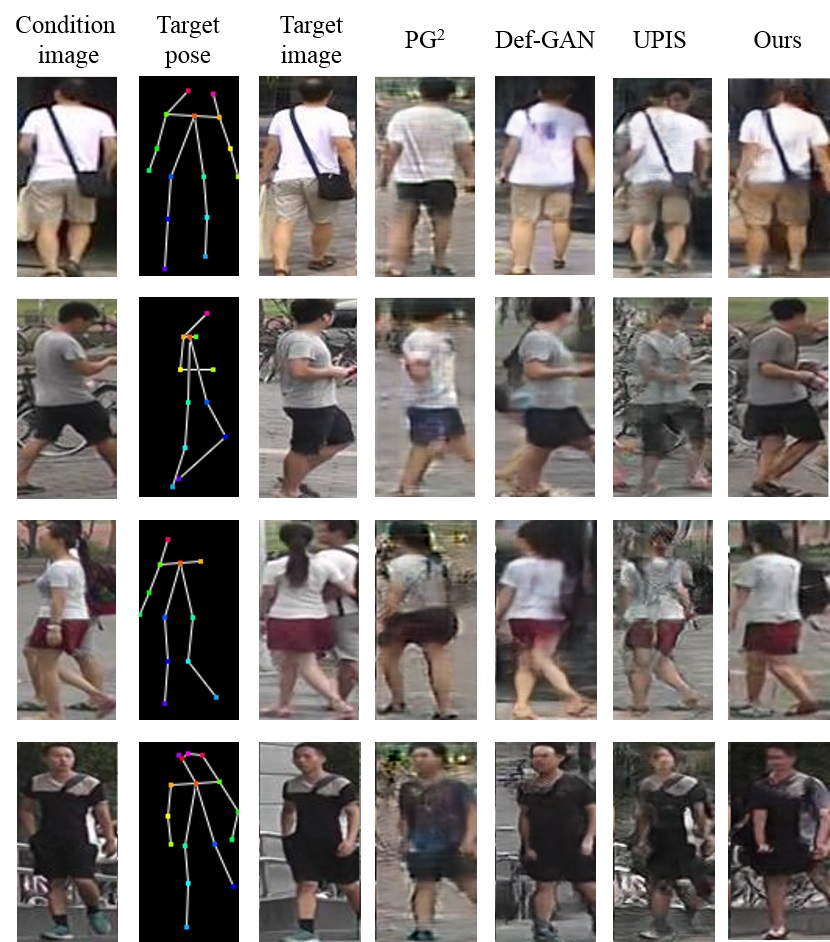}
	\end{center}
    \vspace{-4mm}
	\caption{Example results by different methods (PG${^2}$~\cite{ma2017pose}, Def-GAN~\cite{siarohin2018deformable} and UPIS~\cite{pumarola2018unsupervised}) on Market-1501. Our model generates better body shapes.}
    \vspace{-4mm}
	\label{fig:stoa_market}
\end{figure}
\subsection{Comparison with State-of-the-Arts}

\textbf{Qualitative Comparison.}
In Fig.~\ref{fig:teaser}, Fig.~\ref{fig:stoa_fashion} and Fig.~\ref{fig:stoa_market}, we present the qualitative comparison with three state-of-the-art methods: PG${^2}$~\cite{ma2017pose}, Def-GAN~\cite{siarohin2018deformable} and UPIS~\cite{pumarola2018unsupervised}\footnote{The results for PG$^2$ and Def-GAN are obtained by public models released by their authors, and UPIS are based on our implementation.}. PG${^2}$~\cite{ma2017pose} and Def-GAN~\cite{siarohin2018deformable} are supervised methods that require paired training data. UPIS~\cite{pumarola2018unsupervised} is under the unsupervised setting, which essentially employs CycleGAN~\cite{zhu2017unpaired}. Our model generates more realistic images with higher visual quality and less artifacts. As shown in Fig.~\ref{fig:stoa_fashion}, our method is especially superior in keeping the clothing attributes, including textures and clothing type (the last row). Similarly in Fig.~\ref{fig:stoa_market}, our method better shapes the legs and arms. More generated results can be found in our supplementary.

\textbf{Quantitative Results.}
In Table~\ref{table:quantitative_results}, we use the Inception Score (IS)~\cite{salimans2016improved} and Structural SIMilarity (SSIM)~\cite{wang2004image} for quantitative evaluation. For Market-1501 dataset, to alleviate the influence of background, mask-IS and mask-SSIM are also employed as in~\cite{ma2017pose}, which exclude the background area when computing IS and SSIM. For a fair comparison, we mark the training data requirements for each method. 
Overall, our proposed model achieves the best IS value on both datasets, even compared with supervised methods, which is in agreement with more realistic details and better body shape in our results. Our SSIM score is slightly lower than other methods, which can be explained by the fact that blurry images always achieve higher SSIM but being less photo-realistic, as observed in~\cite{ma2017disentangled,ma2017pose,Johnson2016Perceptual,Shi2016Real}. Limited by space, please refer to our supplementary for user study.

\begin{table*}[t]
	\begin{center}
		\caption{Quantitative results on DeepFashion and Market-1501 datasets (*Based on implementation). }
\vspace{-2mm}
		\label{table:quantitative_results}
		\begin{tabular}{c c c c c c c c}
		\toprule
               &  &\multicolumn{2}{c}{DeepFashion} & \multicolumn{4}{c}{Market-1501}  \\
               \cline{3-8}
		Models & Paired data         & IS    & SSIM  & IS    & SSIM  & mask-IS & mask-SSIM  \\
        \hline
        PG$^2$~\cite{ma2017pose}& Y & 3.090 & 0.762 & 3.460 & 0.253 & 3.435   & 0.792  \\
        Def-GAN~\cite{siarohin2018deformable} & Y & 3.439 & 0.756 & 3.185 & 0.290 & 3.502 & \textbf{0.805}  \\
        V-Unet~\cite{esser2018variational} & N & 3.087 & \textbf{0.786} & 3.214 & \textbf{0.353} & -- & --\\
        BodyROI7~\cite{ma2017disentangled} & N & 3.228 & 0.614 & 3.483 & 0.099 & 3.491 & 0.614  \\
        UPIS~\cite{pumarola2018unsupervised} & N & 2.971 & 0.747 & 3.431* & 0.151* & 3.485* & 0.742*  \\
		\hline
        Baseline & N & 3.140 & 0.698 & 2.776 & 0.157 & 2.814 & 0.714 \\
        TS-Pred & N & 3.201 & 0.724 & 3.462 & 0.180 & 3.546 & 0.740 \\
        TS-GT   & N & 3.350 & 0.740 & 3.472 & 0.200 & 3.675 & 0.749   \\
        E2E(Ours) & N & \textbf{3.441} & 0.736 & \textbf{3.499} & 0.203 &\textbf{ 3.680} & 0.758 \\
        \bottomrule
		\end{tabular}
	\end{center}
\end{table*}
\vspace{-4mm}
\begin{figure*}[htbp] 
	\begin{center}
\vspace{-2mm}
    \subfigure[Results on DeepFashion with different configurations. (Note E2E refines the haircut in the 1st row, sleeve length in the 2nd, arms in the 3rd row, compared with TS-Pred.)]{
       \includegraphics[width=0.82\linewidth]{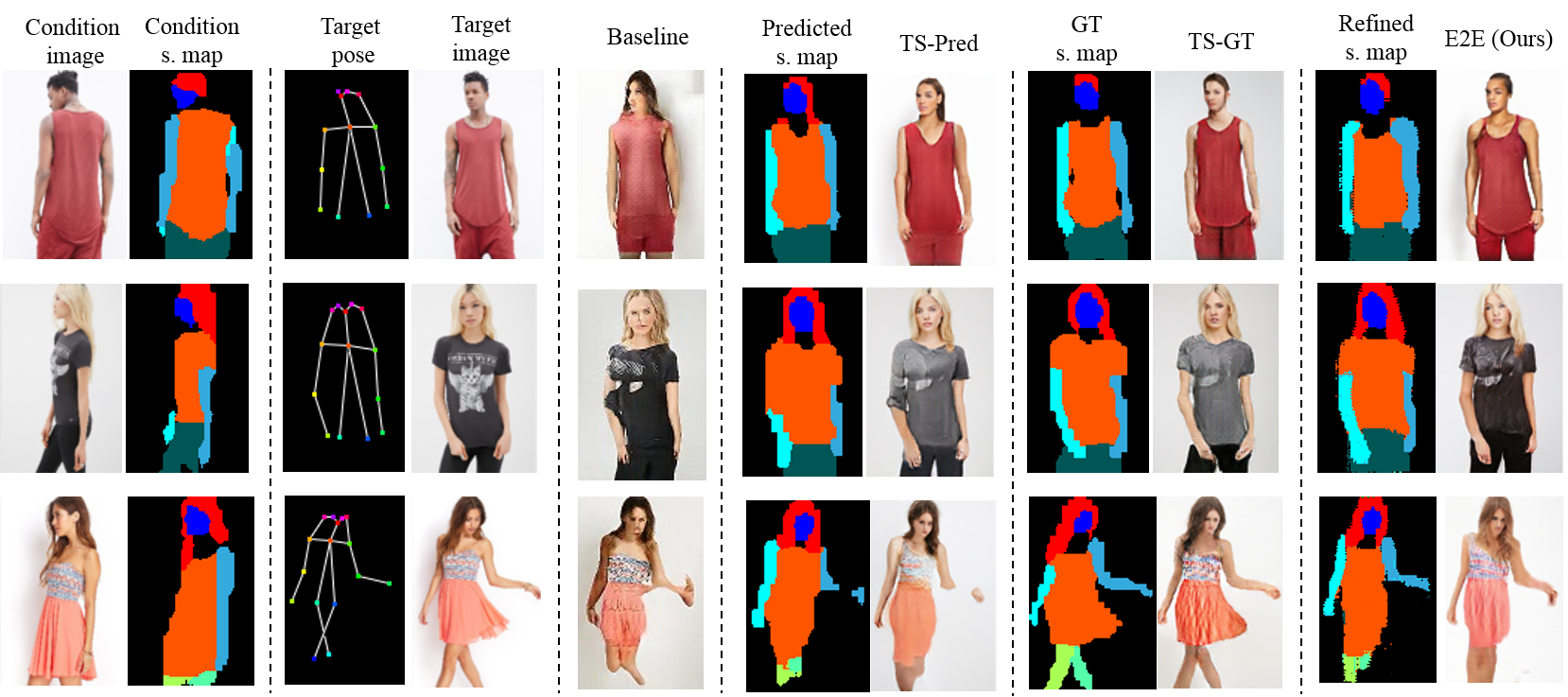}
        \label{fig:ablation_fashion}
    }
\vspace{-2mm}
    \subfigure[Results on Market-1501 with different configurations. (Note E2E refines the body shape in the 1st and 3rd rows, pants length in the 2nd row, compared with TS-Pred.)]{
       \includegraphics[width=0.82\linewidth]{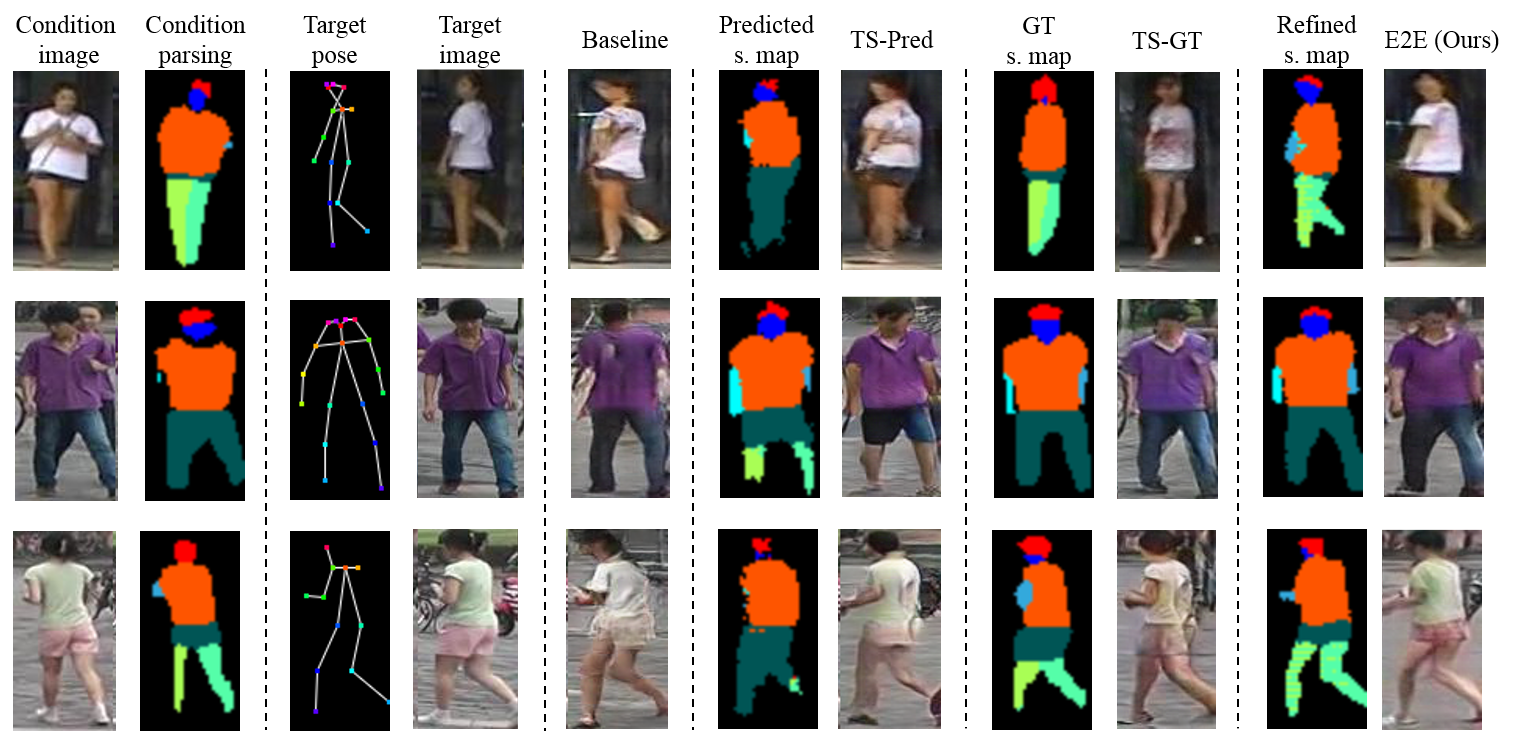}
        \label{fig:ablation_market}
    }
	\end{center}
\vspace{-4mm}
	\caption{Ablation studies on semantic parsing transformation.}
	\label{fig:ablation}
\end{figure*}

\subsection{Ablation Study}

We design the following experiments with different configurations to first evaluate the introduction of semantic information for unpaired person image generation:

$\bullet$ Baseline: our baseline model without the introduction of semantic parsing, the architecture of which is the same as appearance generative network, but without semantic map as input. To keep the style on the output image, we use mask-style loss, which replaces semantic maps with body part masks in Eq.~(\ref{equ:image_style_loss}).

$\bullet$ TS-Pred: The semantic and appearance generative networks are trained independently in a two-stage manner. And we feed the predicted semantic maps into appearance generative network to get the output.

$\bullet$ TS-GT: The networks are trained in two-stage. We regard semantic maps extracted from target images as ground truth, and feed them into appearance generative network to get the output.

$\bullet$ E2E (Ours): jointly training the networks in an end-to-end manner.

Fig.~\ref{fig:ablation} presents the intermediate semantic maps and the corresponding generated images. Table~\ref{table:quantitative_results} further shows the quantitative comparisons. Without the guidance of semantic maps, the network is difficult to handle the shape and appearance at the same time. The introduction of semantic parsing transformation consistently outperforms our baseline. When trained in two-stage, the errors in the predicted semantic maps lead to direct image quality degradation. With end-to-end training, our model is able to refine the semantic map prediction. For example, the haircut and sleeves length in Fig.~\ref{fig:ablation_fashion} are well preserved. For DeepFashion, the end-to-end training strategy leads to comparable results with that using GT semantic maps. For Market-1501, our model (E2E) achieves even higher IS and SSIM values than TS-GT. This is mainly because the human parser~\cite{gong2017look} does not work very well on low-resolution images and many errors exists in the parsing results, as the first row in Fig.~\ref{fig:ablation_market}.

\begin{figure}[t] 
	\begin{center}
		\includegraphics[width=1.0\linewidth]{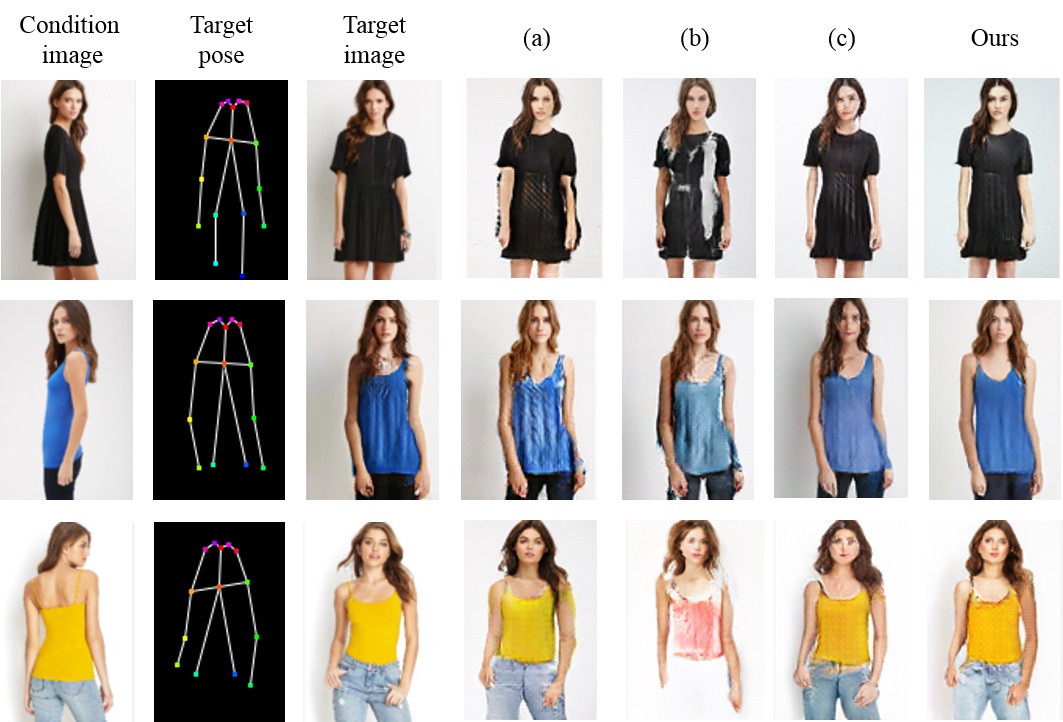}
	\end{center}
    \vspace{-2mm}
	\caption{Analysis for the loss function in appearance generation. (a) Replace $\mathcal L^{sty}_A$ with mask-style loss. (b) Replace $\mathcal L^{sty}_A$ with patch-style loss. (c) Without $\mathcal{L}^{face}_A$. Results of TS-GT with our full loss are in the right.}
    \vspace{-4mm}
	\label{fig:ablation_loss}
\end{figure}

We then analyze the loss functions in the appearance generation as shown in Fig.~\ref{fig:ablation_loss}. We mainly explore the proposed style loss and face adversarial loss, since other losses are indispensable to ensure the cycle consistency. We adopt TS-GT model here to avoid the influence of semantic map prediction. In (a) and (b), we replace the semantic-aware style loss $\mathcal L_A^{sty}$ with mask-style loss and patch-style loss, respectively. Without semantic guidance, both of them lead to dizzy contour. Besides, the adversarial loss for faces effectively helps generate natural faces and improve the visual quality of output images.

\subsection{Applications}
\vspace{-1mm}
Since the appearance generative network essentially learns the texture generation guided by semantic map, it can also be applied on other conditional image generation tasks. Here we show two interesting applications to demonstrate the versatility of our model.

\begin{figure}[t] 
	\begin{center}
        \includegraphics[width=1.0\linewidth]{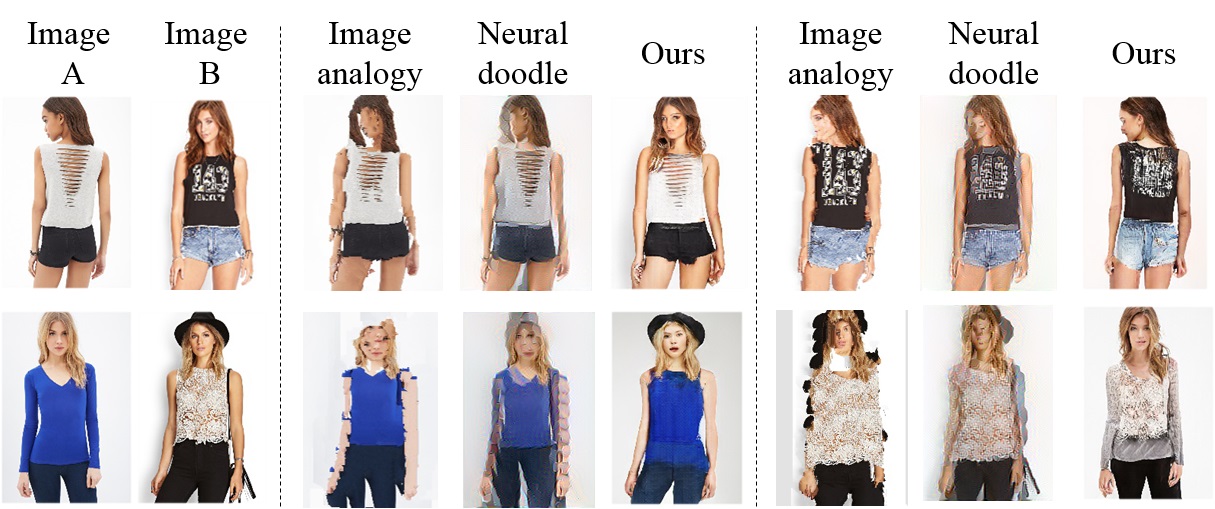}
	\end{center}
\vspace{-3mm}
	\caption{Application for clothing texture transfer. Left: condition and target images. Middle: transfer from A to B. Right: transfer from B to A. We compare our methods with image analogy~\cite{Hertzmann2001Image} and neural doodle~\cite{Champandard2016Semantic}.}
	\label{fig:texture_transfer}
\vspace{-5mm}
\end{figure}

\textbf{Clothing Texture Transfer.}
Given the condition and target images and their semantic parsing results, our appearance generative network is able to achieve clothing texture transfer. The bidirectional transfer results can be viewed in Fig.~\ref{fig:texture_transfer}. Compared with image analogy~\cite{Hertzmann2001Image} and neural doodle~\cite{Champandard2016Semantic}, not only textures are well preserved and transferred accordingly, but also photo-realistic faces are generated automatically.

\begin{figure}[t] 
	\begin{center}
    \includegraphics[width=0.8\linewidth]{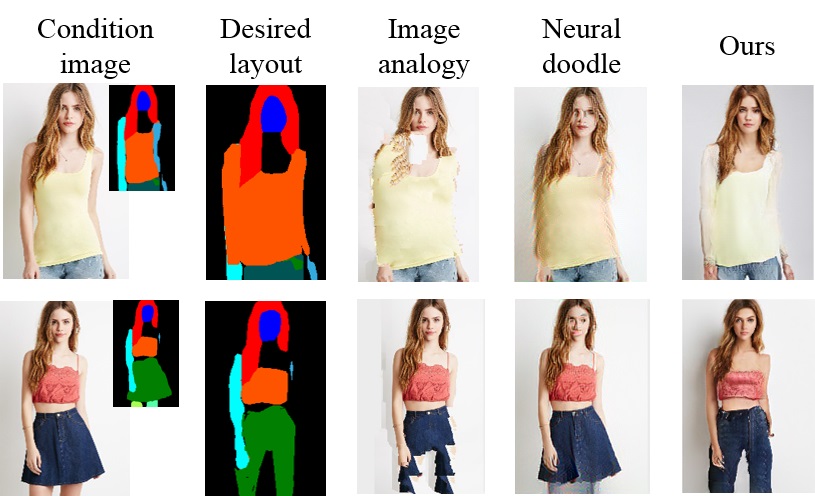}
	\end{center}
\vspace{-3mm}
	\caption{Application for controlled image manipulation. By manually modifying the semantic maps, we can control the image generation in the desired layout.}
\vspace{-3mm}
	\label{fig:controlled_manipulation}
\end{figure}

\textbf{Controlled Image Manipulation.}
By modifying the semantic maps, we generate images in the desired layout. In Fig.~\ref{fig:controlled_manipulation}, we edit the sleeve lengths (top), and change the dress to pants for the girl (bottom). We also compare with image analogy~\cite{Hertzmann2001Image} and neural doodle~\cite{Champandard2016Semantic}.

\subsection{Discussions for Failure Cases}
\begin{figure}[t] 
	\begin{center}
		\includegraphics[width=1.0\linewidth]{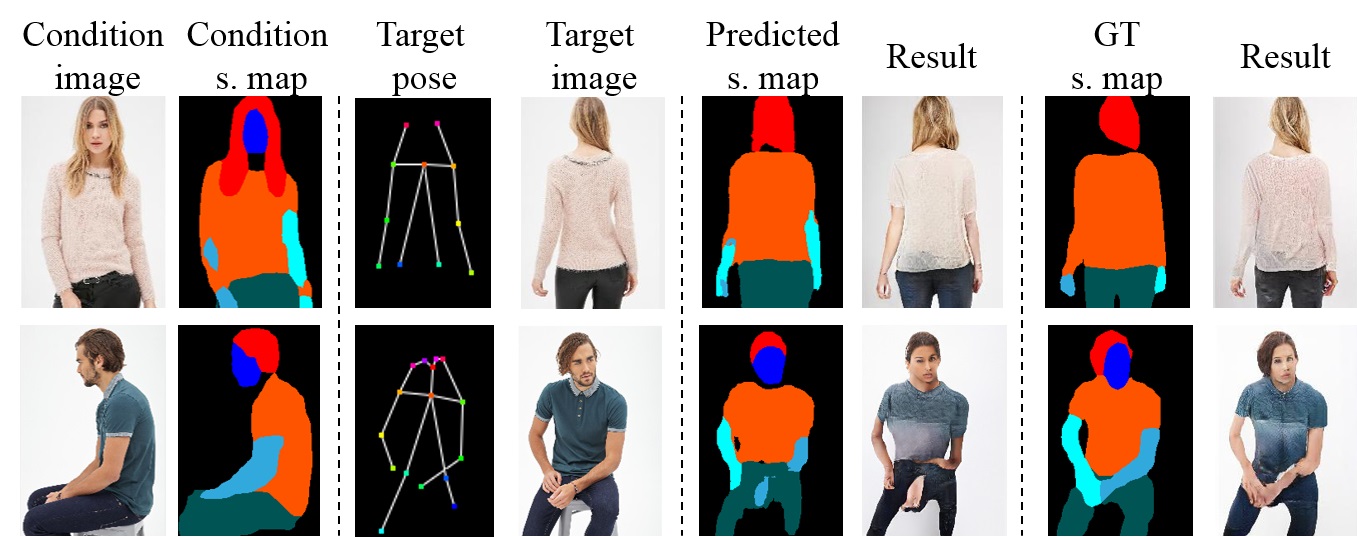}
	\end{center}
\vspace{-5mm}
	\caption{The failure cases in our model.}
\vspace{-5mm}
	\label{fig:failure_case}
\end{figure}

\vspace{-1mm}
Though our model generates appealing results, we show the examples of failure cases in Fig.~\ref{fig:failure_case}. The example in the first row is mainly caused by the error in condition semantic map extracted by the human parser. The semantic generative network is not able to predict the correct semantic map where the arms should be parsed as sleeves. The transformation in the second example is very complicated due to the rare pose, and the generated semantic map is less satisfactory, which leads to unnatural generated images. However, with groundtruth semantic maps, our model still achieves pleasant results. Thus, such failure cases can be probably solved with user interaction.

\section{Conclusion}
In this paper, we propose a framework for unsupervised person image generation. To deal with the complexity of learning a direct mapping under different poses, we decompose the hard task into semantic parsing transformation and appearance generation. We first explicitly predict the semantic map of the desired pose with semantic generative network. Then the appearance generative network synthesizes semantic-aware textures. It is found that end-to-end training the model enables a better semantic map prediction and further final results. We also showed that our model can be applied on clothing texture transfer and controlled image manipulation. However, our model fails when errors exist in the condition semantic map. It would be an interesting future work to train the human parser and person image generation model jointly.

\noindent\textbf{Acknowledgements.} This work was supported by National Natural Science Foundation of China under contract No. 61602463 and No. 61772043, Beijing Natural Science Foundation under contract No. L182002 and No. 4192025.

{\small
\bibliographystyle{ieee_fullname}
\bibliography{CVPR2019_FashionGeneration_arxiv}
}

\end{document}